\documentclass{article}
\usepackage{spconf,amsmath,graphicx}
\usepackage{amsfonts,amssymb}
\usepackage{color}
\usepackage{algorithm}
\usepackage{algpseudocode}
\usepackage{booktabs}
\usepackage{longtable}
\usepackage{multirow}
\usepackage{stfloats}

\title{Self-supervised Guided Hypergraph Feature Propagation for Semi-supervised Classification with Missing Node Features}
\name{Chengxiang Lei, Sichao Fu$^{*, \ddag}$, Yuetian Wang, Wenhao Qiu, Yachen Hu,  Qinmu Peng$^{\ddag}$ and Xinge You\thanks{This work was supported in part by the National Natural Science Foundation of China under Grant 62172177, in part by the Fundamental Research Funds for the Central Universities under Grant 2022JYCXJJ034 and YCJJ202204016. $^{*}$Work was done when Sichao Fu was interning at JD Retail POMC. $^{\ddag}$Sichao Fu and Qinmu Peng are the corresponding authors.}}
\address{School of Electronic Information and Communications, Huazhong University of Science and Technology}

\begin{document}
%
\maketitle
\begin{abstract}
Graph neural networks (GNNs) with missing node features have recently received increasing interest. Such missing node features seriously hurt the performance of the existing GNNs. Some recent methods have been proposed to reconstruct the missing node features by the information propagation among nodes with known and unknown attributes. Although these methods have achieved superior performance, how to exactly exploit the complex data correlations among nodes to reconstruct missing node features is still a great challenge. To solve the above problem, we propose a self-supervised guided hypergraph feature propagation (SGHFP). Specifically, the feature hypergraph is first generated according to the node features with missing information. And then, the reconstructed node features produced by the previous iteration are fed to a two-layer GNNs to construct a pseudo-label hypergraph. Before each iteration, the constructed feature hypergraph and pseudo-label hypergraph are fused effectively, which can better preserve the higher-order data correlations among nodes. After then, we apply the fused hypergraph to the feature propagation for reconstructing missing features. Finally, the reconstructed node features by multi-iteration optimization are applied to the downstream semi-supervised classification task. Extensive experiments demonstrate that the proposed SGHFP outperforms the existing semi-supervised classification with missing node feature methods.
\end{abstract}

 \begin{figure*}[t]
    \centering
    \includegraphics[width=\linewidth]{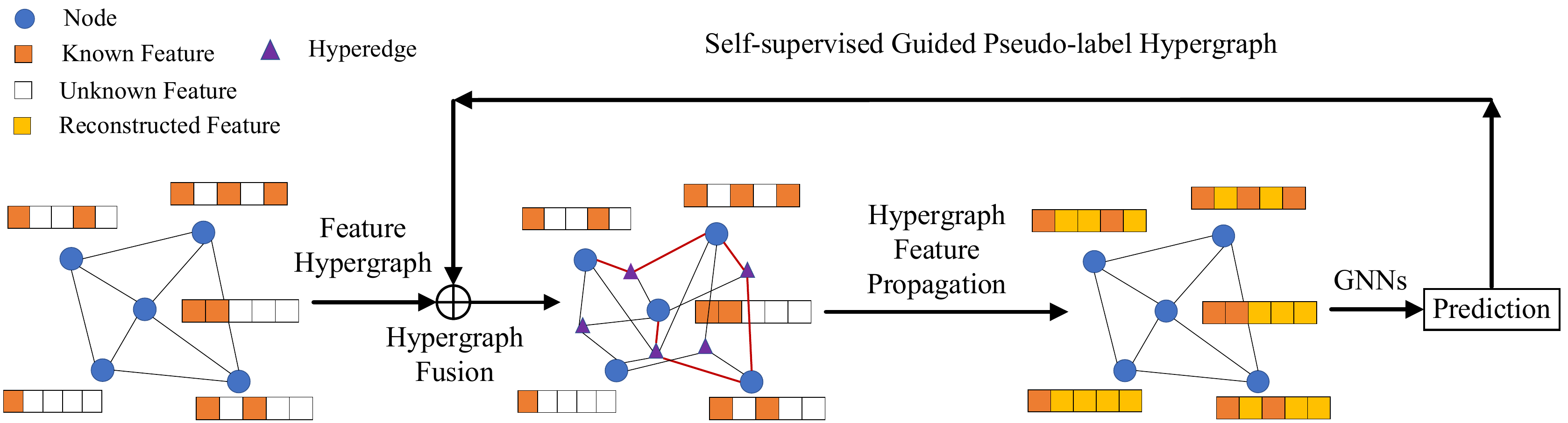}
    \caption{A diagram illustrating our Self-supervised Guided Hypergraph Feature Propagation framework.}
\end{figure*}

\section{Introduction} 
\label{sec:intro}

Graphs are widely applied for many real-world scenarios, such as social networks \cite{hu2022optimal}, citation networks \cite{bayer2022label}, traffic networks \cite{lan2022dstagnn}, molecular networks \cite{stark20223d}.
In recent years, with the rapid development of deep learning, graph neural networks (GNNs) \cite{tiwari2022exploring, fu2022adaptive} have achieved remarkable success in graph-structured data with complex data correlations. GNNs typically operate by a message propagation scheme, where each node propagates its feature representations along the constructed edges. Meanwhile, the feature representation of each node is updated by aggregating the representations from its neighbors.

However, the existing GNNs variants \cite{fu2021semi} typically assume that all node feature attributes are fully observed during the training process. In fact, in many real-world applications, some feature attributes are unobserved due to resource limits or privacy concerns \cite{rossi2021unreasonable}. For example, in social networks, some users are unwilling to provide their ages and gender. Thus, when directly utilizing the existing GNNs variants to deal with the above tasks with missing node features, their performance will have a big deterioration. 

To solve the above issue, many matrix factorization-based methods \cite{liu2019multiple, yoon2018gain} have been proposed to reconstruct missing node features. However, these methods cannot make full use of the topological information between data. Recently, graph signal processing \cite{narang2013signal} provides several methods for interpolating signals on graphs, they generalize Fourier analysis to graphs to reconstruct missing signals on graphs. Such methods are too computationally intensive so they are infeasible for graphs with thousands of nodes. Very recently, several methods extend GNNs to tackle missing node features directly. For example, SAT \cite{chen2020learning} assumes that the structure and feature information on the graph share the same latent space and develops a distribution-matching strategy to reconstruct missing features. GCNMF \cite{taguchi2021graph} adapts GCN \cite{welling2016semi} to graphs with missing features by representing the missing features with a Gaussian mixture model. PaGNN \cite{jiang2020incomplete} develops partial aggregation-based GNNs that only propagate the observed features. FP \cite{rossi2021unreasonable} propagates the known features to the nodes with unknown features iteratively to reconstruct missing features. However, FP only considers pairwise connection relationships between data and FP also assumes that each node has the same influence on all neighbors, which cannot accurately describe the local geometric distribution between data.

In this paper, we propose a self-supervised guided hypergraph feature propagation (SGHFP) for semi-supervised classification with missing node features. Specifically, the feature and pseudo-label hypergraph are first computed according to the node features with missing information and reconstructed node features generated by the previous iteration in turn. Before each iteration, the obtained feature and pseudo-label hypergraph are further fused into an effective hypergraph. Compared to the single feature or pseudo-label hypergraph, the fused hypergraph can better describe the complex high-order structure information between data. Following, the fused hypergraph and FP are combined to reconstruct missing features in each iteration. After multi-iterations optimization, the reconstructed node features can be applied to downstream semi-supervised classification tasks. To validate the effectiveness of SGHFP, we conduct extensive experiments on four benchmarks. Experiment results demonstrate that our approach outperforms many state-of-the-art methods. The main contributions are summarized as follows:

\begin{itemize}
 
\item[$\bullet$] Compared with the traditional graph, the proposed hypergraph can simultaneously utilize the higher-order correlations from feature and pseudo-label hypergraphs to update the inaccurate connection relationships. 

\item[$\bullet$] Our proposed SGHFP is an independent module, which can combine any GNNs variants for any graph representation learning tasks.

\item[$\bullet$] Extensive experiments show that the proposed SGHFP outperforms many existing semi-supervised classifications with missing node feature methods.

\end{itemize}


\section{Self-supervised Guided Hypergraph Feature Propagation}
\label{sec:format}

\subsection{Problem Definition and Notation Description}
\label{ssec:subhead}

Give a simple graph $\mathcal{G} = (V, E)$, where $V = \{v_i | i = 1, \dots n\}$ is the set of nodes, and $E = \{ e_{ij} |  i = 1, \dots n, j = 1, \dots n\}$ is the set of edges. Let $A \in \mathbb{R}{^{n \times n}}$ denotes the adjacency relationship matrix, where $A_{ij} = 1$ if $e_{ij} \in E$, and $A_{ij} = 0$ if $e_{ij} \not\in E$. $X \in \mathbb{R}{^{n \times d}}$ denote the nodes features matrix, where $d$ is the dimension of node features. 

Different from a simple graph, a hyperedge can connect two or more nodes. Let $H = (\tilde{V}, \tilde{E}, W)$ denotes a hypergraph, where $\mathbf{W}$ denotes the weight matrix of all hyperedges, and $H \in \mathbb{R}^{|\tilde{V}| \times |\tilde{E}|}$ is a incidence matrix, i.e.  

\begin{equation}
	h(\tilde{v},\tilde{e}) = \begin{cases}
	1, \quad $if$ \quad \tilde{v} \in \tilde{e} \\
	0, \quad $otherwise$
		   \end{cases}
\end{equation}

For a node $\tilde{v} \in \tilde{V}$, its degree is defined as $d(\tilde{v}) = \Sigma_{\tilde{e} \in \tilde{E}}w(\tilde{e})h(\tilde{v},\tilde{e})$. For a hyperedge $\tilde{e} \in \tilde{E}$, its degree is defined as $\delta(\tilde{e}) = \Sigma_{\tilde{v} \in \tilde{V}}h(\tilde{v},\tilde{e})$. Let $\mathbf{D}_e$ denote all hyperedges' degree matrix and $\mathbf{D}_v$ denote all nodes' degree matrix. In addition, we let $\mathbf{\Theta}=\mathbf{D}_v^{-1/2}\mathbf{HWD}_e^{-1}\mathbf{H}^\top\mathbf{D}_v^{-1/2}$ and $\mathbf{\Delta} = \mathbf{I} - \mathbf{\Theta}$, where $\mathbf{\Delta}$ is called as hypergraph Laplacian.

$\tilde{V}_{k} \subseteq \tilde{V}$ denotes the set of nodes where the features are known, and $\tilde{V}_u = V_k^c = \tilde{V} \backslash \tilde{V}_k$ denote the unknown ones. In this paper, the problem we focus on is how to better reconstruct the unknown features $\mathrm{\tilde{x}}_u$, given the known features $\mathrm{\tilde{x}}_k$ and the graph structure $\mathcal{G}$.

\subsection{Hypergraph Fusion}
\label{sssec:subsubhead}

To accurately describe the local geometric distribution among nodes, We first generate a feature hypergraph $G_f$ according to the node features with missing information. Each time one node is selected as the centroid, and all its neighbors are linked as a hyperedge. Second, the reconstructed node features produced by the previous iteration are fed to a two-layer GNNs to construct a pseudo-label hypergraph $G_{pl}$. 

Denote $v_a$ and $v_b$ are two nodes, from \cite{wang2020unifying}, we can know that the label influence of $v_a$ on $v_b$ equals to the to the cumulative normalized feature influence of $v_a$ on $v_b$ after k iterations of propagation:  
\begin{equation}\label{eql:2}
\mathbb{E}[I_l(v_a,v_b;k)] = \Sigma_{j=1}^k\tilde{I}_f(v_a,v_b,j). 
\end{equation}

Equation \ref{eql:2} shows that pseudo-label hypergraph is beneficial to increase the intra-class feature influence. To highlight the accuracy of connection relationships, the constructed feature hypergraph $G_f$ and pseudo-label hypergraph $G_{pl}$ are further combined to generate a fused hypergraph before each iteration. To reduce complexity, we fuse the two graphs into a sparse matrix. $G_f$ is used as the indices of specified elements, and $G_{pl}$ is used as the corresponding values. As shown in Fig.1, hyperedges connecting two nodes of the same class are bold so that features can be more easily propagated among nodes along hyperedges with stronger connections.   

\subsection{Hypergraph Feature Propagation}
\label{sssec:subsubhead}
Similar to FP \cite{rossi2021unreasonable}, we reconstruct the unknown node features $\mathrm{\tilde{x}}_u$ through interpolation that minimizes \textit{Dirichlet energy}:  $\ell(\mathrm{x},\mathcal{G}) = \frac{1}{2}x^\top  \mathbf{\Delta}\mathrm{x} = \frac{1}{2}\Sigma_{ij}\theta_{ij}(x_i-x_j)^2$, where $\theta_{ij} $ are the individual entries of the normalized incidence matrix $\mathbf{\Theta}$. Dirichlet energy, which represents how much a function changes in a certain area, is widely used as a smoothness criterion.

For the convenience of derivation, the node feature matrix $\tilde{\mathrm{x}}$ is split into two sub-matrices. H and $\mathbf{\Delta}$  can be divided into four sub-matrices.
\begin{equation}
    \mathrm{\tilde{x}} = \begin{bmatrix} \mathrm{\tilde{x}}_k \\ \mathrm{\tilde{x}}_u  \end{bmatrix}  \mathbf{H} = \begin{bmatrix} \mathbf{H}_{kk} & \mathbf{H}_{ku} \\ \mathbf{H}_{uk} & \mathbf{H}_{uu} \end{bmatrix} \mathbf{\Delta} = \begin{bmatrix} \mathbf{\Delta}_{kk} & \mathbf{\Delta}_{ku} \\ \mathbf{\Delta}_{uk} & \mathbf{\Delta}_{uu} \end{bmatrix}.
\end{equation}
Let $\dot{\mathrm{x}}(t) = - \nabla \ell(\tilde{\mathrm{x}}(t))$ denotes the associated \textit{gradient flow}, and the known features $\tilde{\mathrm{x}}_k =\tilde{\mathrm{x}}_k(t)$ is the boundary condition. Therefore, the solution at the missing nodes: $\tilde{\mathrm{x}}_u = \lim_{n \to \infty}\tilde{\mathrm{x}}_u(t)$ is the interpolation.
From \cite{rossi2021unreasonable}, we can get a diffusion equation:
{
\small
\begin{equation}
\label{equation:3}
\setlength{\arraycolsep}{1pt}
\begin{bmatrix} 
\dot{\mathrm{x}}_k(t)\\ \dot{\mathrm{x}}_u(t)
\end{bmatrix} \!\!=\!\!-\!
\begin{bmatrix}  0 & 0 \\ \mathbf{\Delta}_{uk} & \mathbf{\Delta}_{uu} \end{bmatrix} \! \!
\begin{bmatrix} \tilde{\mathrm{x}}_k \\ \tilde{\mathrm{x}}_u(t) \end{bmatrix}\!\!=\! \!-\!\begin{bmatrix} 0 \\ \mathbf{\Delta}_{uk}\tilde{\mathrm{x}}_k + \mathbf{\Delta}_{uu}\tilde{\mathrm{x}}_u(t) 
\end{bmatrix}. 
\end{equation}
}
and its solution in an iterative scheme: 
\begin{equation}
\label{equation:4}
\tilde{\mathrm{x}}^{(k+1)} = \begin{bmatrix} \mathbf{I} & 0 \\ \mathbf{\Theta}_{uk} & \mathbf{\Theta}_{uu} \end{bmatrix} \tilde{\mathrm{x}}^{(k)}.
\end{equation}

\begin{algorithm}[h]
  \caption{SGHFP}
  \label{algorithm1}
  \begin{algorithmic}[1]
    \State \textbf{Input}:feature vector x, graph structure $G$, train epochs T
    \For{$\mathrm{x}=1$ $\rightarrow$ T}
        \State  construct $G_f$ and $G_{pl}$ by x and $G$
        \State $\mathbf{\Theta} \gets fuse(G_f, G_{pl})$
        \State  $\mathrm{x} \gets \mathbf{\Theta}  \mathrm{x} $ \Comment{Propagate features}
        \State  $\mathbf{x}_k \gets \mathbf{y}_k$ \Comment{Reset known features}
    \EndFor
  \end{algorithmic}
\end{algorithm}

The update procedure in equation 4 is equivalent to the following two steps. First, the feature vector x is multiplied by the propagation matrix $\mathbf{\Theta}$. Second, the known features are reset to their original true values. This update procedure provides an iterative algorithm to reconstruct the unknown node features, as shown in Algorithm \ref{algorithm1}. Specifically, the feature hypergraph $ G_f$ and pseudo-label hypergraph $G_{pl}$ are fused to generate a propagation matrix $\mathbf{\Theta}$ (in the first iteration, only feature hypergraph $G_f$ is used). At each iteration, features are propagated among nodes by the propagation matrix $\mathbf{\Theta}$. After that, we clamp the known features by resetting them to their original true values.

\section{Experiment}
\label{sec:experiment}

\subsection{Datasets}
\label{ssec:subhead}
We fed the reconstructed node features into many downstream semi-supervised node classification tasks, their classification performance can reflect the accuracy of feature reconstruction intuitively. In this paper, we evaluate our proposed SGHFP on four benchmark datasets including Cora, CiteSeer, PubMed \cite{sen2008collective}, and Photo (Amazon) \cite{wang2020microsoft}.

\begin{table}[H]
\caption{Dataset statistics.}
\centering
\begin{tabular}{ccccc} 
    \hline
    Dataset  & Nodes & Edges  & Features & Classes  \\ 
    \hline
    Cora     & 2485  & 5069   & 1433     & 7        \\
    Citeseer & 2120  & 3679   & 3703     & 6        \\
    PubMed   & 19717 & 44324  & 500      & 3        \\
    Photo    & 7487  & 119043 & 745      & 8        \\
    \hline
    \end{tabular}
    \label{table:datasets}
\end{table}

\subsection{Experimental Setup}
\label{ssec:subhead}

In all experiments, We randomly select 20 nodes per class as the training set, 1500 nodes for validation, and the rest for testing. In this paper, we use a two-layers GCN with a dropout rate of 0.5 as the downstream classifier. The Adam optimizer \cite{kingma2015adam} with a learning rate of 0.005 is introduced to optimize the model parameters. Empirically, hypergraph feature propagation diffuses the features over 50 iterations. For all the baselines, the hyperparameters are the same as mentioned in the respective papers \cite{rossi2021unreasonable,taguchi2021graph,jiang2020incomplete,xiaojin2002learning,dwivedi2020benchmarking}.

\begin{table}[htp]
    \caption{Classification performance of our proposed SGHFP and FP under different rates of missing features. The best results are highlighted.}
    \centering
    \resizebox{\linewidth}{!}{
    \begin{tabular}{l|l|cccc} 
    \hline
    \multirow{2}{*}{Dataset}  & \multirow{2}{*}{Method} & \multicolumn{4}{c}{Missing Node Features Rate}                                        \\
                              &                         & 0 & 50\%             & 90\%             & 99\%              \\ 
    \hline
    
    \multirow{2}{*}{Cora}     & FP                      & 80.39\%       & 79.7\% (-0.86\%)  & 79.77\% 
     (-0.77\%) & 78.22\% (-2.70\%)  \\
                              & SGHFP                    & \textbf{81.23\%}       & \textbf{80.9\%} (-0.41\%)  & \textbf{80.41\% }
     (-1.01\%) & \textbf{79.41}\% (-2.25\%)  \\ 
    \hline
    
    \multirow{2}{*}{Citeseer} & FP                      & \textbf{67.48}\%       & 65.74\% (-2.57\%) & 65.57\% 
     (-2.82\%) & 65.4\% (-3.08\%)   \\
                              & SGHFP                    & 67.45\%       & \textbf{66.86\%} (-0.87\%) & \textbf{66.76\%} 
     (-1.02\%) & \textbf{66.5\%} (-1.41\%)   \\ 
    \hline
    
    \multirow{2}{*}{PubMed}   & FP                      & 77.36\%       & 76.68\% (-0.89\%) & 75.85\% 
     (-1.96\%) & 74.29\% (-3.97\%)  \\
                              & SGHFP                    & \textbf{77.45}\%       & \textbf{77.15\%} (-0.37\%) & \textbf{76.63\%} 
     (-1.06\%) & \textbf{75.17\%} (-2.94\%)  \\ 
    \hline
    
    \multirow{2}{*}{Photo}    & FP                      & 91.73\%       & 91.29\% (-0.48\%) & 89.48\% 
     (-2.46\%) & 87.73\% (-4.36\%)  \\
                              & SGHFP                    & \textbf{92.08\%}      & \textbf{91.46\%} (-0.67\%) & \textbf{90.04\%} 
     (-2.22\%) & \textbf{88.40\%} (-3.99\%)  \\
    \hline
    \end{tabular}
    }
    \label{table:compare FP}
\end{table}

\subsection{Results and Discussion}
\label{ssec:subhead}

\subsubsection{Comparison with State-of-the-art Methods}
\label{sssec:subsubhead}

We report the mean classification accuracy with 10 runs for all methods. Each run has a different mask of missing features. Table 2 illustrates the classification performance of our proposed SGHFP and FP (combined with a downstream GCN model) under full features, 50\%, 90\%, and 99\% missing node features. In addition, we also report the relative drop rate when features are partially missing compared to all features that are present. Table \ref{table:compare FP} shows that the proposed SGHFP outperforms FP under any missing rate and also has a lower relative drop rate. On average, SGHFP with 99\% missing features loses only 2.67\% of relative accuracy compared to the same GNNs model with no missing features.

\begin{table*}[ht]
    \caption{Ablation experiments on Cora, Citseer, PubMed, and Photo dataset with 99\% missing node features. }
    \centering
    \resizebox{\linewidth}{!}{
        \begin{tabular}{l|cccccccccccc} 
        \hline
        Module  & \multicolumn{3}{c}{Cora}    & \multicolumn{3}{c}{Citeseer} & \multicolumn{3}{c}{PubMed}      & \multicolumn{3}{c}{Photo}  \\ 
        \hline
        FP    &    \checkmark     &   \checkmark      & \checkmark       &   \checkmark      &  \checkmark   & \checkmark        &    \checkmark     &   \checkmark      & \checkmark      & \checkmark        &    \checkmark     &   \checkmark  \\
        FH              &         & \checkmark       & \checkmark       &         & \checkmark       & \checkmark        &         & \checkmark       & \checkmark        &         & \checkmark       & \checkmark        \\ 
        SGPH                    &         &         & \checkmark       &         &         & \checkmark        &         &         & \checkmark    &         &         & \checkmark     \\
        \hline
        Accuracy                & 78.22$\pm$0.32 & 78.77$\pm$0.24 & 79.41$\pm$0.21 & 65.40$\pm$0.54 & 65.92$\pm$0.28 & 66.50$\pm$0.11  & 74.29$\pm$0.55 & 74.56$\pm$0.36 & 75.17$\pm$0.10 & 87.73$\pm$0.27 & 88.02$\pm$0.18 & 88.40$\pm$0.10  \\
        \hline
        \end{tabular}
        }
        \label{table:ablation}
\end{table*}

\begin{figure*}[htp]
\begin{minipage}[b]{.24\linewidth}
  \centering
  \centerline{\includegraphics[width=\linewidth]{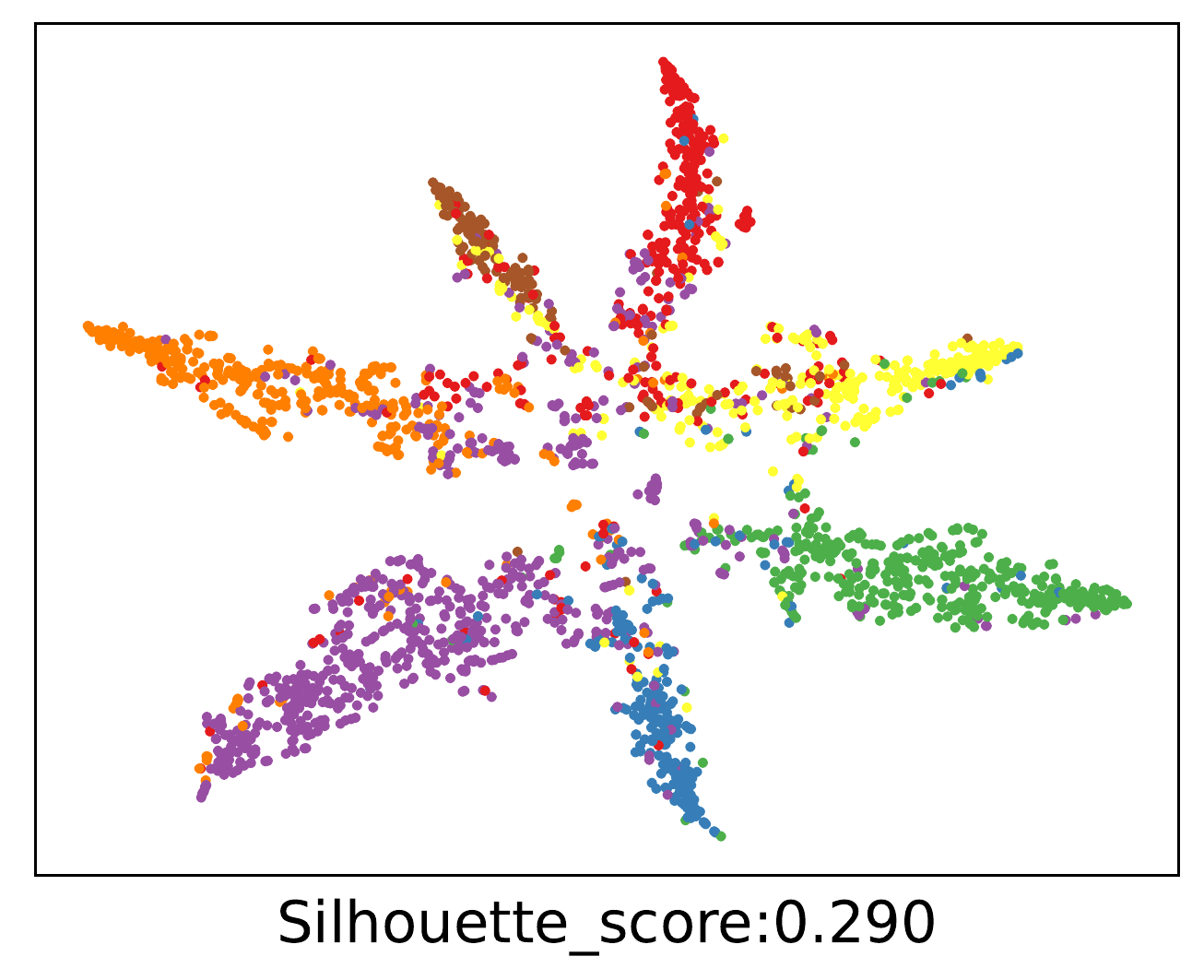}}
  \centerline{(a) FP on Cora}\medskip
\end{minipage}
\hfill
\begin{minipage}[b]{0.24\linewidth}
  \centering
  \includegraphics[width=\linewidth]{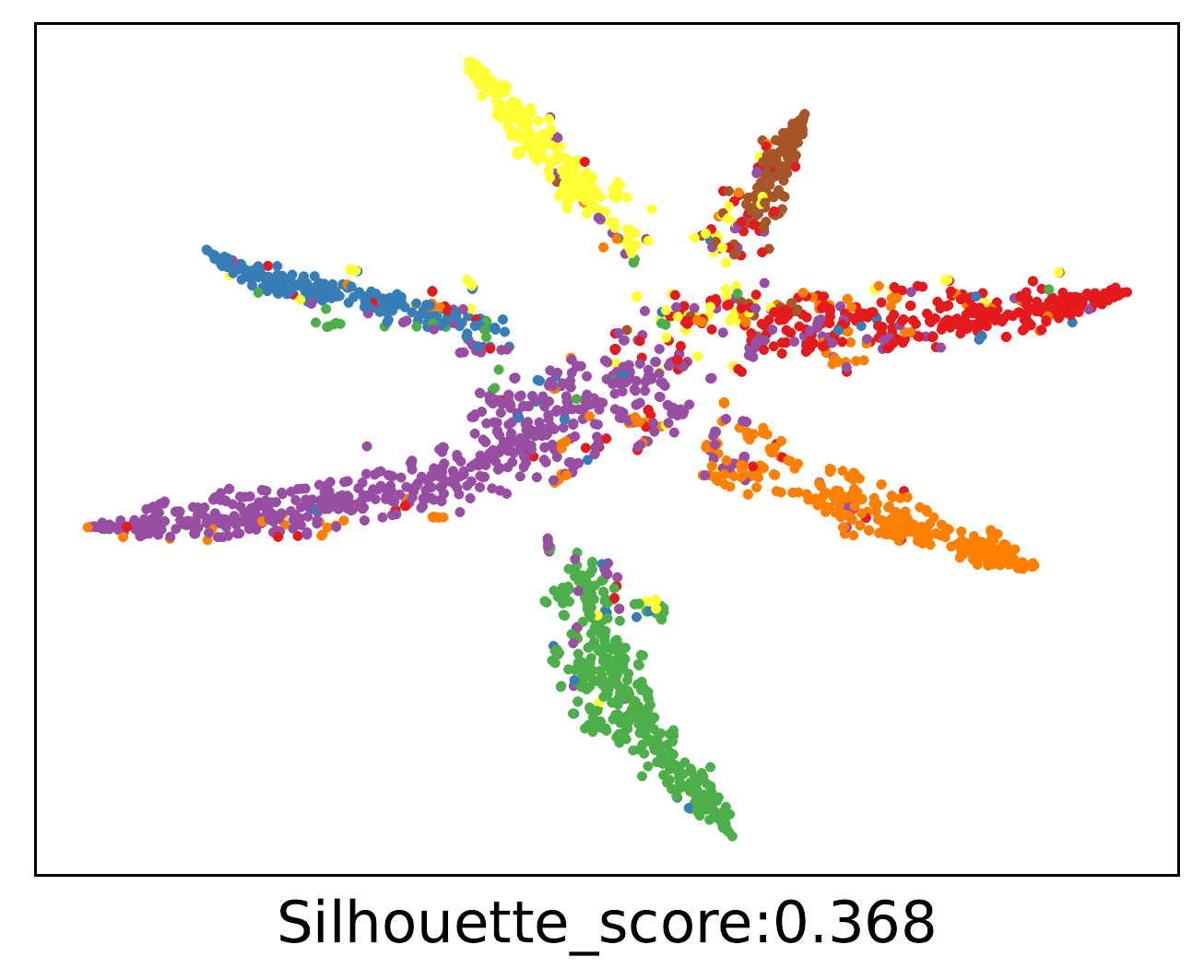}
  \centerline{(b) SGHFP on Cora}\medskip
\end{minipage}
\begin{minipage}[b]{.24\linewidth}
  \centering
  \centerline{\includegraphics[width=\linewidth]{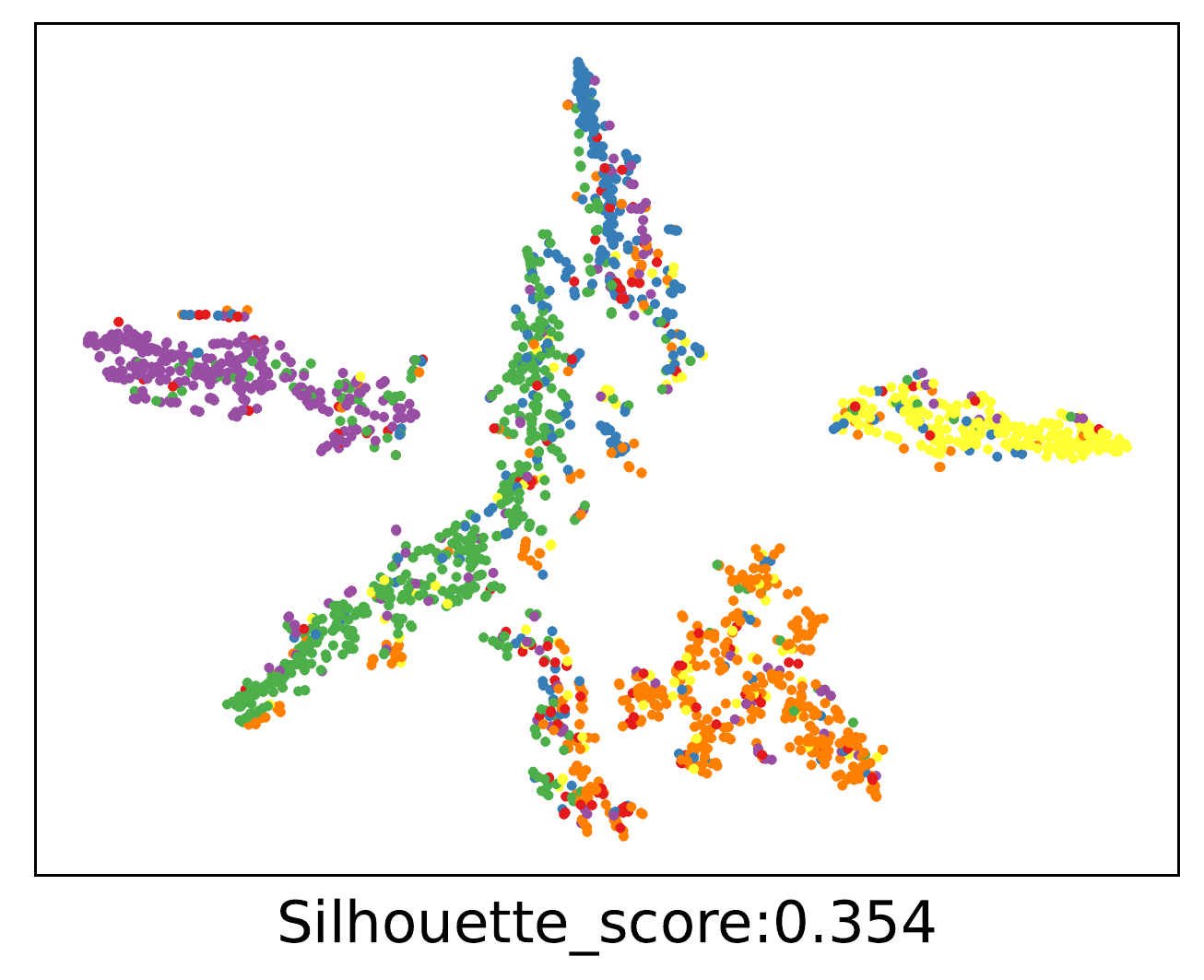}}
  \centerline{(c) FP on Citeseer}\medskip
\end{minipage}
\hfill
\begin{minipage}[b]{0.24\linewidth}
  \centering
  \centerline{\includegraphics[width=\linewidth]{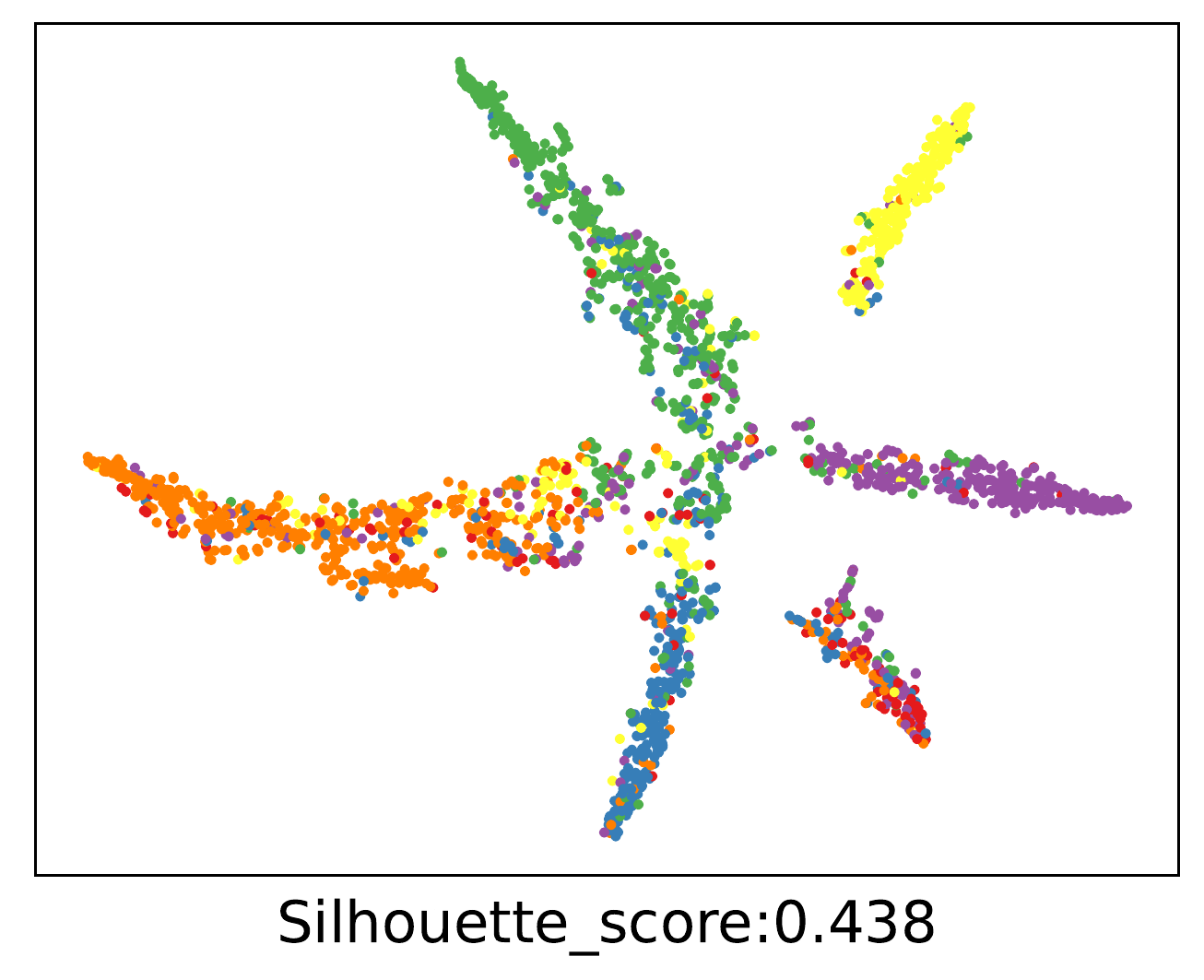}}
  \centerline{(d) SGHFP on Citeseer}\medskip
\end{minipage}
\caption{The t-SNE visualization and Silhouette score of the node embeddings with reconstructing features. Each color represents one class.}
\label{fig:2}
\end{figure*}

\begin{table}[H]
    \caption{Classification performance of GCNMF, PaGNN, LP, GPE and SGHFP under 99\% of features missing. The best results are highlighted.}
    \centering
    \resizebox{\linewidth}{!}{
    \begin{tabular}{cccccc}
    \toprule
     Dataset  & GCNMF      & PaGNN      & LP    & GPE    & SGHFP \\
    \midrule
     Cora     & 34.54$\pm$2.07 & 58.03$\pm$0.57 & 74.68$\pm$0.36  & 76.33$\pm$0.26 & \textbf{79.41}$\pm$0.21\\
    Citeseer & 30.65$\pm$1.12 & 46.02$\pm$0.58 & 64.60$\pm$0.40  & 65.87$\pm$0.37 & \textbf{66.50}$\pm$0.11 \\
    PubMed   & 39.80$\pm$0.25 & 54.25$\pm$0.70 & 73.81$\pm$0.56  & 73.70$\pm$0.29 & \textbf{75.17}$\pm$0.10\\
    Photo    & 29.64$\pm$2.78 & 85.41$\pm$0.28 & 83.45$\pm$0.94  & 83.45$\pm$0.26 & \textbf{88.40}$\pm$0.10\\
    \bottomrule
    \end{tabular}
    }
    \label{table:compare others}
\end{table}

In this part, we compare the proposed SGHFP with many state-of-the-art GNNs methods when 99\% of the node features are missing. We additionally compare to feature-agnostic baselines: Label Propagation (LP) \cite{xiaojin2002learning}, which only exploits the structure information of graphs by propagating labels iteratively, and Graph Positional Encodings (GPE) \cite{dwivedi2020benchmarking}, which treats the first $k$ eigenvector matrices of the Laplacian as node features. Table \ref{table:compare others} shows that our proposed SGHFP outperforms the existing methods on all experimental datasets. For example, GCNMF and PaGNN have a large drop in relative accuracy when the feature missing rate is high. In comparison, SGHFP has only a 2.67\% drop.
 
\subsubsection{Ablation experiments}
\label{sssec:subsubhead}

In this part, we investigate the impact of the self-supervised guided pseudo-label hypergraph (SGPH) and feature hypergraph (FH) module for feature reconstruction on Cora, Citseer, PubMed, and Photo datasets. Table \ref{table:ablation} shows that our proposed SGPH and FH modules all improve the classification performance of downstream tasks with missing node features. For example, on the Cora dataset, FP with the FH module obtains gains of 0.55\%. FP with SGPH and FH module (SGHFP) achieve 1.19\% improvements in comparison to FP.

\subsubsection{t-SNE visualization}
\label{sssec:subsubhead}

To better demonstrate that the proposed SGHFP can better reconstruct the missing features, we use t-SNE and Silhouette scores to visualize the embedding of graphs with reconstructed features in 2-D space on Cora and Citeseer datasets. Nodes in the same class are expected to be clustered together and have higher Silhouette scores. Fig.\ref{fig:2} shows that our proposed SGHFP can separate different categories and the nodes in the same class are clustered more compactly. All experiments are performed with a 90\% missing rate of node features.

\section{Conclusion}
\label{sec:illust}

In this paper, we present a novel approach for semi-supervised classification with missing node features. The feature hypergraph and pseudo-label hypergraph are constructed to describe the local geometric distribution between data in turns. Then, a fused hypergraph generated by an effective strategy is further applied to the feature propagation model for reconstructing the missing features. Experimental results on several datasets demonstrate that the proposed SGHFP is useful to reconstruct the missing features and also outperforms many existing state-of-the-art methods. While SGHFP is designed for homogeneous graphs, it does not perform well on heterogeneous graphs. Learning heterogeneous feature-missing graphs with a more general learnable diffusion matrix could also be an interesting problem.

\clearpage

\bibliographystyle{IEEEbib}
\bibliography{strings}

\end{document}